\documentclass{article}
\pdfoutput = 1
\usepackage[square, comma, sort&compress, numbers]{natbib}

\usepackage[preprint]{neurips_2022}

\usepackage[utf8]{inputenc} 
\usepackage[T1]{fontenc}    
\usepackage{url}            
\usepackage{booktabs}       
\usepackage{amsfonts}       
\usepackage{nicefrac}       
\usepackage{microtype}      
\usepackage{xcolor}         
\usepackage{enumitem}
\usepackage{graphicx}
\usepackage{bigstrut}
\usepackage{makecell}
\usepackage{booktabs} 
\usepackage{array}
\usepackage{amsmath}
\usepackage{caption}
\usepackage{float}

\title{MiNL: Micro-images based Neural Representation for Light Fields}

\author{
	Hanxin Zhu \\
	University of Science and Technology of China\\
	\texttt{hanxinzhu@mail.ustc.edu.cn } \\
	\And
	Henan Wang \\
	University of Science and Technology of China \\
	\texttt{henanwang@mail.ustc.edu.cn } \\
	\AND
	Zhibo Chen \\
	University of Science and Technology of China \\
	\texttt{chenzhibo@ustc.edu.cn} \\
}
\begin{document}
\maketitle
	\begin{abstract}
		Traditional representations for light fields can be separated into two types: explicit representation and implicit representation. Unlike explicit representation that represents light fields as Sub-Aperture Images (SAIs) based arrays or Micro-Images (MIs) based lenslet images, implicit representation treats light fields as neural networks, which is inherently a continuous representation in contrast to discrete explicit representation. However, at present almost all the implicit representations for light fields utilize SAIs to train an MLP to learn a pixel-wise mapping from 4D spatial-angular coordinate to pixel colors, which is neither compact nor of low complexity. Instead, in this paper we propose MiNL, a novel MI-wise implicit neural representation for light fields that train an MLP + CNN to learn a mapping from 2D MI coordinates to MI colors. Given the micro-image's coordinate, MiNL outputs the corresponding micro-image's RGB values. Light field encoding in MiNL is just training a neural network to regress the micro-images and the decoding process is a simple feedforward operation. Compared with common pixel-wise implicit representation, MiNL is more compact and efficient that has faster decoding speed (\textbf{$\times$80$\sim$180} speed-up) as well as better visual quality (\textbf{1$\sim$4dB} PSNR improvement on average). With such a representation, all information of light fields are stored in parameters of neural networks, which can realize several light field-related tasks at the same time. For example, compared with mainstream light field compression methods that have complex processing pipeline, our proposed method transform the light field compression task into model compression task and can achieve comparable performance with state-of-the-art methods through a simple neural network training, with about \textbf{1$\sim$2dB} PSNR improvement over HEVC/H.265 at the same bit rate. In addition to light field compression, MiNL can also be generalized to light field denoising, with more than \textbf{7dB} PSNR improvement over original noisy light field images for different kinds of noises.
	\end{abstract}

	\section{Introduction}
	Derived from the plenoptic function, a light ray can be parameterized by its intersections with two parallel planes, namely the spatial plane $(x, y)$ and the angular plane $(u, v)$. Thus light field contains four-dimensional (4-D) information. With the additional dimensions compared to traditional 2-D images, light field images can bring us more applications such as refocusing and view synthesis, which can lead to more immersive experience in scenarios such as virtual reality (VR).
	
	Normally light field images are captured by lenslet cameras. Different from conventional cameras, a micro-lens array is inserted between the main lens and the sensor plane. Each micro-lens captures a low resolution portion of the scene, namely a micro-image (MI) or macro-pixel. A light field image from Lytro camera is composed of 625×434 micro-images with 15×15 pixels inside each micro-image. By extracting pixels in the same position of each micro-image, we get a series of sub-aperature images (SAI), which can be regarded as photos captured by cameras from different viewing angles.
	
	Whether MI-based or SAI-based representations, both of them share a common drawback: the enormous data. As mentioned above, a LF image from Lytro camera has 625×434 spatial resolution and 15×15 angular resolution. No matter one treats it as a huge image of size (9375, 6510) or 225 images of size (625, 434), it's extremely inconvenient for transmission or storage. As a result, a more efficient representation for light field is severely demanded.

	\begin{figure}
		\centering
		\setlength{\belowcaptionskip}{-0.5cm}
		\includegraphics[height=9.5cm]{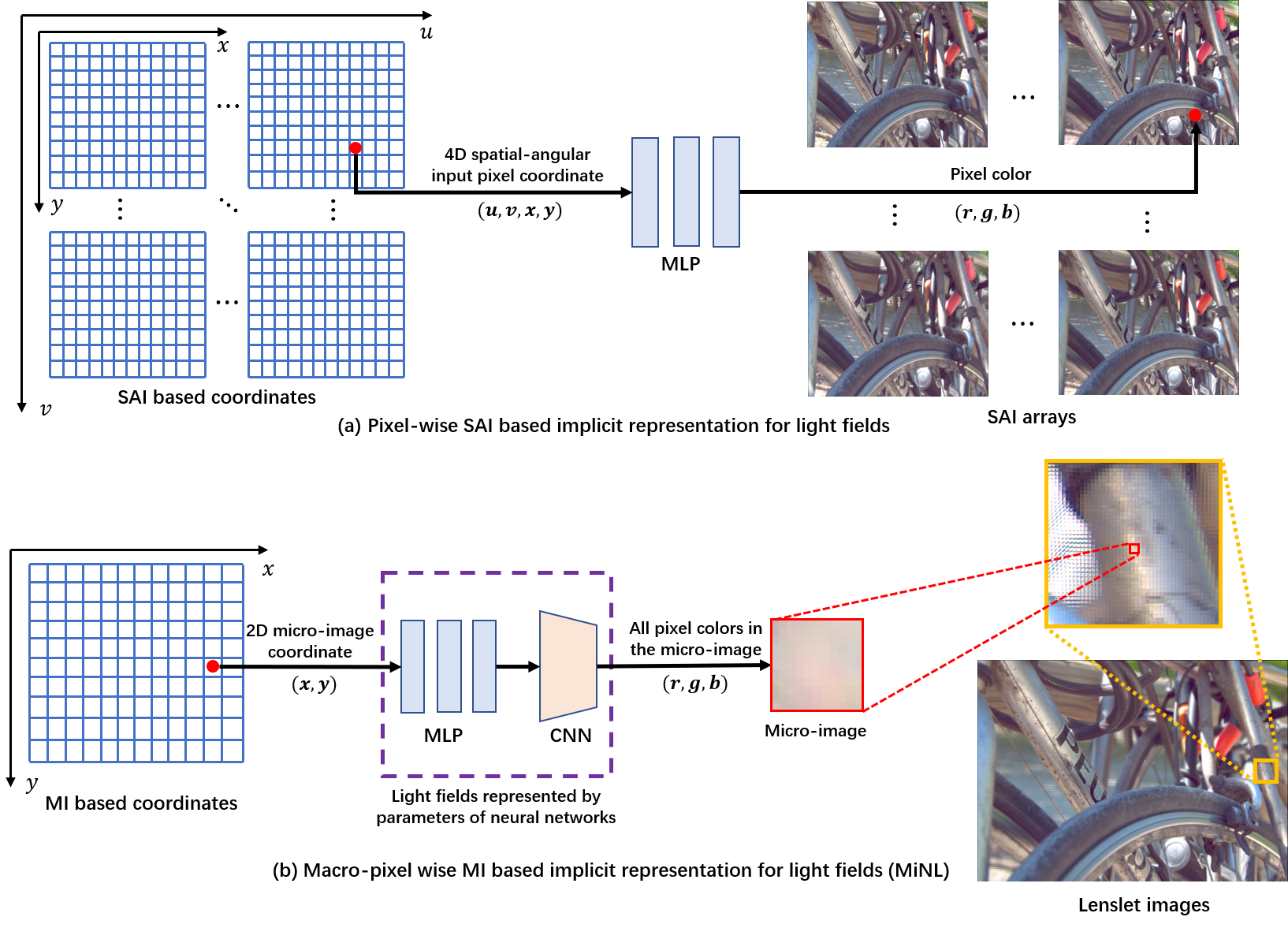}
		\caption{(a) Common pixel-wise implicit representation for light fields which takes 4D spatial-angular coordinate as input on the basis of SAIs     (b) Our proposed method, MiNL, which takes 2D micro-image coordinate as input on the basis of MIs.}
		\label{fig 1}
	\end{figure}
		
\begin{figure}
		\centering
		\setlength{\belowcaptionskip}{-0.2cm}
		\includegraphics[height=7.5cm]{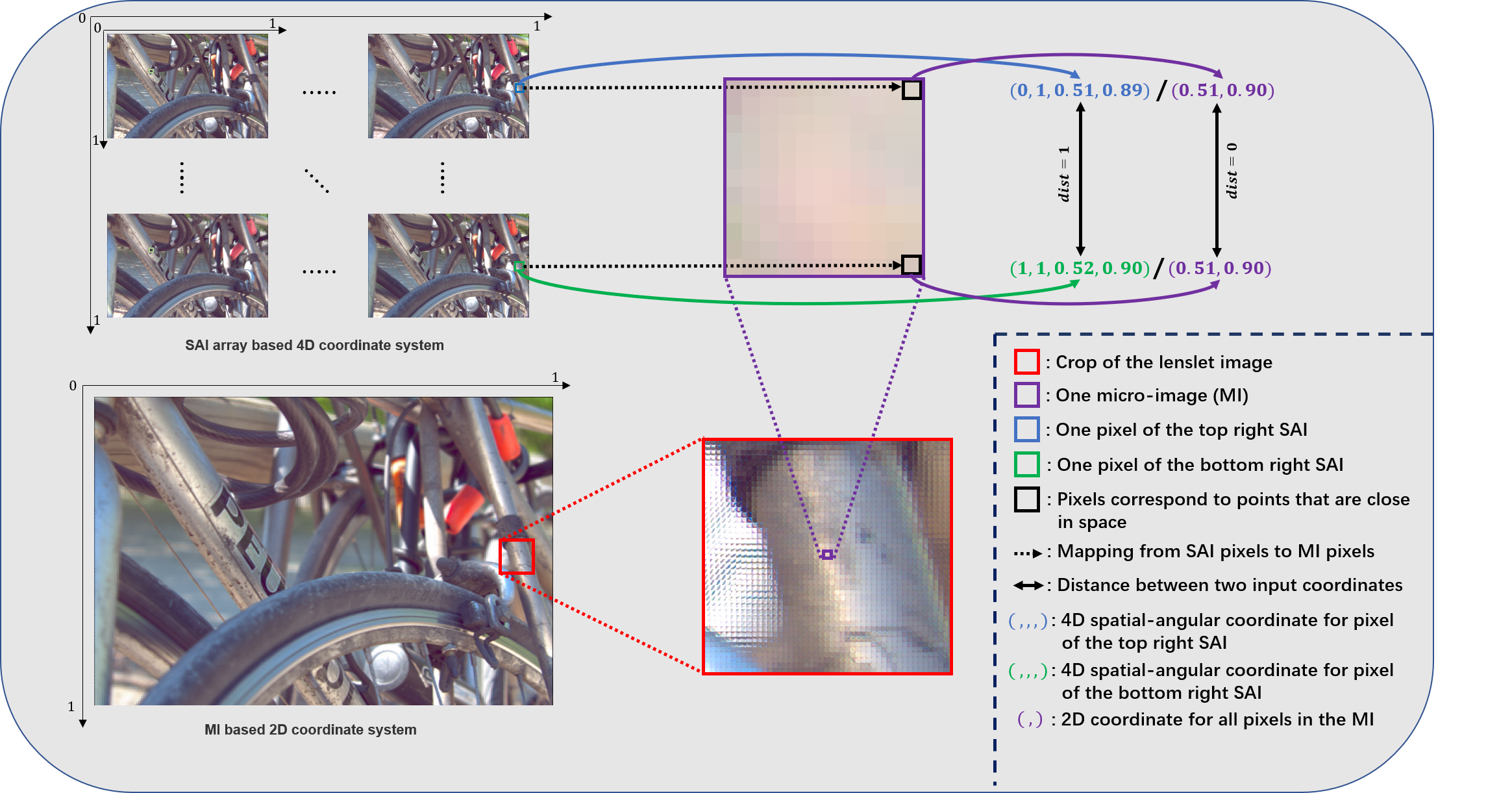}
		\caption{Differences of input coordinates for SAI based pixel-wise implicit representation and MI based MI-wise implicit representation}
		\label{fig 2}
	\end{figure}

	To solve the above-mentioned problem, we propose MiNL, a novel neural representation for light fields. As is shown in Figure \ref{fig 1} , different from most common and naive pixel-wise implicit representations for light fields that utilize sub-aperture images to construct a mapping from 4D spatial-angular coordinates to pixel colors \cite{bemana2020x,chandramouli2021light,feng2021signet,dupont2021coin}, MiNL is a MI-wise implicit representation which take advantage of micro-images to learn a mapping from 2D micro-image coordinates to micro-image colors. Compared with the traditional implicit representation, MiNL owns two significant advantages: low complexity and higher (or more compact) representation ability. Fistly, on account that one micro-image usually has hundreds of pixels, one feedforward process of MiNL is equivalent to hundreds of feedforward processes of pixel wise implicit representation, and thus the decoding speed of MiNL is much faster ($\times$80$\sim$180 speed-up). Secondly, just like the pixels in the blue box and the green box shown in the upper left corner of Figure \ref{fig 2} , for pixels of different views correspond to spatial points that are close to each other with similar RGB values, their input coordinates deserve to be adjacent as well. In other words, distances of input coordinates for these spatially adjacent points should be as small as possible. However, this is not the case for pixel-wise SAI based implicit representation, where two spatially adjacent points with almost the same colors can have quite different input coordinates, as demonstrated by the blue coordinate and the green coordinate in the upper right corner of Figure \ref{fig 2} , which will lead to more model capacity to remember these meaningless differences. In contrast, for MI based implicit representation, i.e. MiNL, all pixels in one micro-image are inherently spatially adjacent points which have the same input coordinates, as shown in the purple box and the purple coordinates in Figure \ref{fig 2} . Such a characteristic promises the superiority of our proposed method, where MiNL can achieve more compact representation for light fields and obtain higher visual quality under circumstances of the same model capacity compared with mostly adopted SAI based pixel-wise implicit representation nowadays.
	
	Morever, our proposed neural representation for light fields also has a great advantage over traditional light field representations, \textit{i.e.} SAIs or MIs, and is able to realize several light field-related tasks through a simple neural network training. For example, to achieve light field compression, classical coding frameworks such as HEVC/H.265 ususally treat all of the sub-aperture images as pseudo video sequences, followed by several processing modules such as key frames selection, residual calculation, discrete cosine transform, mode selection, predictive coding and so on to obtain the compressed light field images \cite{7574673,7137669,linearapprox}. In addition to these hand-crafted coding method, view synthesis based methods are also popular for light field compression, where a few of key view SAIs are selected to be coded in the coding process and the rest unselected SAIs are synthesised in the decoding process \cite{huang2019light,huang2020low,huang2018view,jia2018light}. All of these methods depend on a complex processing pipeline and the decoding operation is quite complicated as well. In contrast, MiNL leverages the model parameters of neural networks to store light field images, and thus transform the light field compression task into model compression task. By means of several simple model compression techniques such as model pruning and model quantization, experimental results have demonstrated that MiNL can achieve comparable performance with state-of-the-art light field compression methods, with about 1$\sim$2dB PSNR improvement over HEVC/H.265 at the same bit rate.
	
	In addition to light field compression, MiNL also shows its remarkable ability for light field denoising. Given several noisy light field images, one can obtain the denoised images simply by virtue of the direct neural network training as mentioned above, without any other operations. Extensive experiments are conducted for different light field images with different types of noises, the result shows more than 7dB PSNR improvement between original noisy images and the denosied images, which surpass some traditional filter-based denoising methods.
	
	The main contributions of this paper can be summarized as follows:
	\begin{itemize}[leftmargin=*]
	\item We propose a novel micro-image based neural representation for light fields, \textit{i.e.} MiNL. MiNL leverages parameters of neural networks to store information of light fields, which is able to realize several light field related tasks through just a simple neural network training. As far as we know, MiNL is the first work that utilize micro-images instead of sub-aperture images to construct implicit neural representations for light fields.
	
	\item We discovered that micro-images based implicit representation is much more efficient than SAI- based implicit representation for representing light fields. Under condition of the same model capacity, MiNL owns the advantages of lower complexity and higher (or more compact) representation ability, promising the superiority of our proposed method.
	
	\item Compared with common SAI based pixel-wise implicit representations for light fields, MiNL is a MI-wise implicit representation that can achieve faster decoding speed (\textbf{$\times$80$\sim$180} speed up) and better visual qualities (\textbf{1$\sim$4dB} PSNR improvement).
	 
	\item MiNL transforms the light field compression task into model compression task. By means of several model compression techniques, MiNL can achieve comparable performance with state-of-the-art  light field compression methods, with more than \textbf{1$\sim$2dB} PSNR improvement over HEVC/H.265. 
	
	\item Given the noisy light field images, MiNL can obtain the denoised ouput through a simple neural network training without any other operations. The results demonstarte more than \textbf{7dB} denosing improvement in PSNR over original noisy images for different kinds of noises.
	\end{itemize}
	
	\section{Related work}
	
	\subsection{Coordinated-based implicit neural representation}
	Coordinated-based implicit neural representation has gained more and more attention since the success of \cite{mildenhall2020nerf}. In contrast to traditional discrete explicit representations such as mesh, point cloud and voxels, implicit representation is actually an continuous representation which is quite memeory-efficient. As a rule, coordinated-based implicit neural representation is realized by training a neural network to learn a mapping from input coordinates to scene properties such as pixel colors \cite{chen2021nerv,park2021nerfies,li2021mine}, volume density \cite{jang2021codenerf,mildenhall2020nerf,xu2022point} and signed diatance \cite{park2019deepsdf}. For implicit representations of light fields, almost all the methods adopt a similar scheme that make use of SAIs to construct a mapping from 4D spatial-angular coordinates to pixel RGB values \cite{bemana2020x,chandramouli2021light,feng2021signet}, which is neither compact nor efficient. Instead, in this paper, we propose a method based on MIs to realize a MI-wise implicit neural representation that learns a function from 2D MI coordinates to colors of all pixles in the MI, \textit{i.e.} MiNL. Experimental results demonstrated the superiority of MiNL from the aspect of lower complexity and higher representation ability.
	
	\subsection{Light Field Compression}
	Due to the vast size, various compression solutions have been proposed to exploit redundancies of light field images. Here we divide them into lenslet-based methods and view-based methods.
	
	\paragraph{Lenslet image compression methods}
	As Figure \ref{fig 2} shows, a lenslet image contains a number of micro-images, forming repetitive patterns. Conti et al.\cite{CONTI201659} proposes a self-similarity (SS) compensated prediction to explore the correlations in MIs. Li et al.\cite{7574673,7137669} propose a method to predict MIs using a scheme similar to HEVC inter-frame prediction. Liu et al.\cite{8793171} propose a content-based LF image-compression method with
	Gaussian process regression to improve the compression efficiency and accelerate the prediction procedure. Inspired by end-to-end image compression networks, Tong et al.\cite{tong2022sadn} propose a end-to-end spatial-angular-decorrelated network (SADN) to take advantage of both spatial or angular consistency in lenslet images.
	
	\paragraph{SAI compression methods}
	SAI representation of light field can be viewed as an array of 2-D images. Thus it's reasonable to organize these images in a specific order and treat them as a pseudo video sequence (PVS). Vieria et al.\cite{vieira2015data} organize SAIs in a PVS with different scan orders and encodes it with HEVC. Liu et al.\cite{liu2016pseudo} propose a 2-D hierarchical structure which divides SAIs into different coding layers according to their positions, then these SAIs are encoded by JEM encoder. Besides, view synthesis methods encodes only a few key SAIs and recovers the rest at the decoder side. Zhao et al.\cite{linearapprox} pick a sparse set of SAIs and encodes them with HEVC. The rest SAIs are synthesized by a linear approximation model using the key SAIs. Several deep learning based methods \cite{bakir2018light,zhao2018light,jia2018light} are also proposed to utilize the CNN architecture to recover the discarded SAIs.
	
	In this paper we propose a new approach to achieve the light field compression task. Instead of designing a complex architecture to compress the explicit form of light field images, we adopt an implicit form (\textit{i.e.} MiNL) to represent light field images and then compress the model parameters directly. Combining a relatively simple network architecture and an efficient model compression scheme, our method can obtain better compression performance.
	
	\subsection{Model Compression}
	As mentioned above, in the form of implicit representation, the light field compression task can be converted into a model compression task. LeCun et al.\cite{lecun1989optimal} have shown that not all network parameters are necessary. Removing unimportant weights can shrink model size and reduce computational complexity, making it possible to deploy neural networks on resource-constrained devices.  By compressing the implicit model, we can further reduce the bitrate while maintaining the image quality. Existing model compression schemes can be categorized into pruning and quantization \cite{han2015learning,wen2016learning,li2016pruning,chen2018shallowing,liu2020dynamic,lee2021layeradaptive}, low-rank approximation \cite{zhang2015efficient,tulloch2017high}, knowledge distillation \cite{hinton2015distilling,romero2014fitnets,chen2015net2net,gou2021knowledge} and so on. In this paper, we take advantage of model pruning, model quantization and entroy coding to realize model compression.
	
	\section{Method}
	\subsection{MI-wise implicit neural representation for lenslet image}
\begin{figure}
		\centering
		\includegraphics[height=7.5cm]{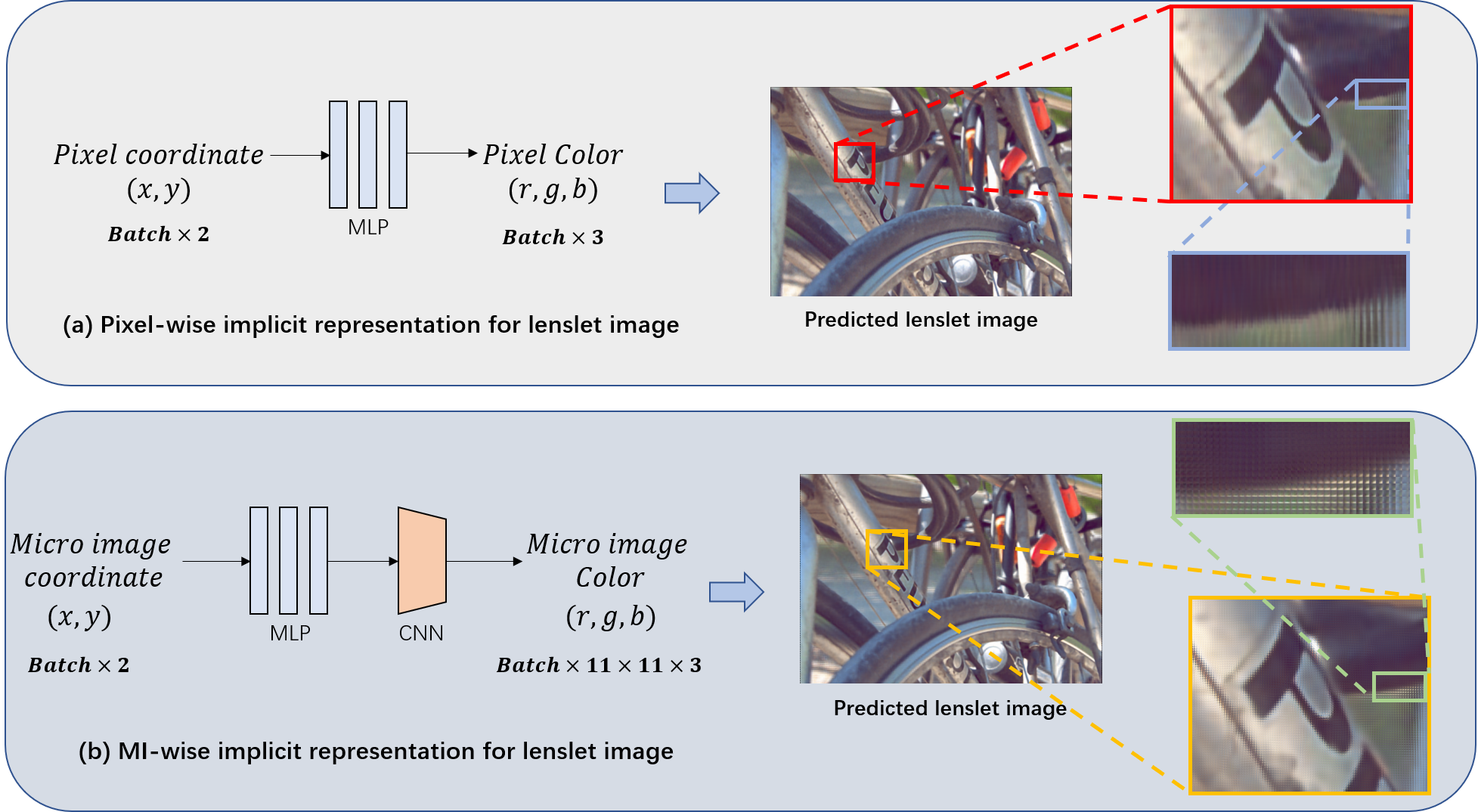}
		\caption{{(a)} Results of pixel-wise implicit representation for lenslet image that have severe leakage  {(b)} Results of MI-wise implicit representation for lenslet image that are leakage-free }
		\label{fig 3}
	\end{figure}

	As is illustrated above, in this paper we choose to leverage MIs to construct an implicit representation for light fields. Considering a lenslet image consists of 625$\times$434 micro-images with 11$\times$11 pixels inside each of them, we can treat it as one giant image composed of pixels whose resolution is 6875$\times$4774 or one relative small image whose resolution is 625$\times$434 composed of micro-images. For the former one, the implicit representation is still pixel-wise that learns a mapping from pixel coordinate to pixel colors, which will be quite time-consuming in the decoding process just like SAI-based pixel-wise implicit representation. Besides, because implicit representation tends to smooth the image, such a pixel-wise method will blur boundaries between different micro-images and lead to severe leakages from adjacent pixels, as shown by the blue box in the upper half of Figure \ref{fig 3} . Instead, for the latter one, \textit{i.e.} MiNL, implicit representation is realized by learning a mapping from the micro-image coordinate to the micro-image colors. Such a MI-wise implicit representation promises a faster decoding speed and will be leakage-free since micro-images are directly predicted rather than pixel by pixel, which is able to preserve the boundaries between different micro-images, as shown by the green box in the lower half of Figure \ref{fig 3} .
	
	\paragraph{Position encoding}
	As is demonstrated by \cite{mildenhall2020nerf}, a direct training from raw input coordinates to scene propertities will lead to quite poor results, where the whole image is excessively smoothed out. As a result, to obtain high frequency detailed information, the raw input coordinates need to be mapped to a high-dimensional space before being fed into the neural network. In this paper, we choose the so-called Position Encoding operation \cite{mildenhall2020nerf} which is based on the Fourier base
	\begin{equation}
		\centering
		\textbf{D}=(\sin(2^0\pi\textbf{d}),\cos(2^0\pi\textbf{d}),\sin(2^1\pi\textbf{d}),\cos(2^1\pi\textbf{d}),\cdots,\sin(2^{L-1}\pi\textbf{d}),\cos(2^{L-1}\pi\textbf{d})),\label{pe}
	\end{equation}
	where $\textbf{d}$ represents the raw 2D input coordinate that is normalized between $[0,1]$, $\textbf{D}$ represents the embedded high-dimensional vector, $L$ is a hyperparameter.
	
	\paragraph{MiNL architecture}
	As is shown in Figure \ref{fig 1} and Figure \ref{fig 3} , MiNL takes the 2D MI coordinate as input and ouputs the corresponding colors of all pixels in the MI. Different from most common pixel-wise implicit representation that utilize one MLP to store scene information, in this paper we propose to make use of MLP + CNN in order to match the dimension of MI. To facilitate training of the neural network, a simple but important trick is that the activation functions for all layers of MLP and CNN are $\textbf{sine}$ function, which is demonstrated by \cite{sitzmann2020implicit} that such a scheme will help improve the speed and robustness of the training process.
	
	\paragraph{Loss function}
	In this paper, we adopt the following loss function to train our model, which is a combination of l2 loss and regularization term:
	\begin{equation}
		\centering
		Loss=\frac{1}{B}\sum_{i=0}^{B}{||M_i^{pre}}-M_i^{gt}||_2+\alpha||\theta||_1,\label{loss}
	\end{equation}
	where $B$ is the number of one batch, $M_i^{pre}$ is the $i$-th predicted micro-image by MiNL, $M_i^{gt}$ is the $i$-th ground truth micro-image, $\theta$ are model parameters and $\alpha$ is a hyperparameter that used to balance the distortion term and the regularization term. 
	
	\subsection{Model compression}
	In this section, we introduce the model compression schemes used in MiNL. We adopt the model compression procedure introduced in \cite{chen2021nerv}, which includes model pruning, model quantization and entropy coding.
	
	\paragraph{Pruning} Pruning is an efficient model compression method which can increase model sparsity by eliminating unnecessary weights. Here we adopt the pruning method proposed in \cite{lee2021layeradaptive}. It first calculates the layer-adaptive magnitude-based pruning (LAMP) score of each weight, which is a rescaled version of the weight magnitude, followed by a global pruning based on LAMP score
	\begin{equation}
		{\hat{w}}_i=\left\{
		\begin{aligned}
			w_{i} & , & \quad{\rm if}\ {\rm score}(w_{i})\geq {\rm score}_{\rm thres} \\
			0 & , & \quad{\rm otherwise}
		\end{aligned}
		\right.
	\end{equation}
	where $w_{i}$ is the $i$-th original model parameter, ${\hat{w}}_i$ is the $i$-th pruned model parameter, ${\rm score}(w_{i})$ is the LAMP score of the $i$-th original model parameter, ${\rm score}_{\rm thres}$ is the LAMP score threshold value for all model parameters.

	\paragraph{Quantization} After model pruning, quantization procedure is executed to further reduce the bits needed for network parameters. A network parameter $\delta_i$ (usually FP-32) is mapped to a $b$-bit integer
	\begin{equation}
		Q(\delta_i) = {\rm round}(\frac{\delta_{i}-\delta_{\min}}{S}), \quad S=\frac{\delta_{\max}-\delta_{\min}}{2^b}
	\end{equation}
	where 'round' is rounding the value to the closest integer, $S$ the scale factor, $\delta_{\max}$ and $\delta_{\min}$ the max and min value for the parameter tensor $\delta$. When decoding, parameter $\delta_i$ can recovered by
	\begin{equation}
		\hat{\delta_{i}}=Q(\delta_i)*S+\delta_{\min}
	\end{equation}
	
	\paragraph{Entropy Coding} After quantization, Huffman Coding\cite{huffman1952method} is adopted as entropy coding model to generate the final bitstream. Huffman Coding distributes different number of bits to parameter values based on their overall frequency and generates a codebook. Different from the processes above, this one is lossless and can recover the quantized parameters without any reconstruction loss.
	
    \begin{figure}
    	\centering 
    	\hspace{-0.7cm}\includegraphics[height=5.5cm]{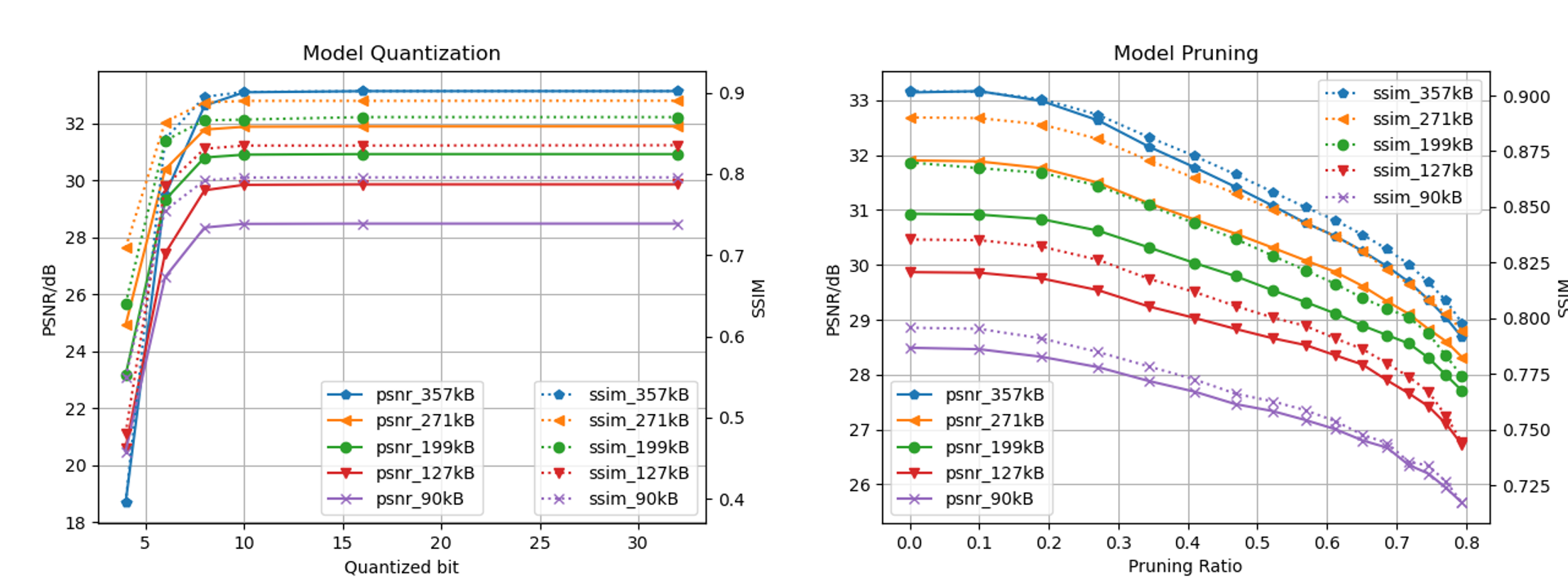}
    	\caption{Model quantization and model pruning for models of different sizes}
    	\label{Quantization and pruning}
    \end{figure}
    \begin{table}[htbp]
    	\centering 
    	\caption{Comparison of MiNL with other implicit neural representation for light fields}
    	\begin{tabular}{c|ccccccc}
    		\hline\label{table1}
    		Method  & \multicolumn{1}{p{3.04em}}{SIREN\newline{}\makecell[c]{\cite{sitzmann2020implicit}}} & \multicolumn{1}{p{3.04em}}{SIGNET\newline{}\makecell[c]{\cite{feng2021signet}}} & \multicolumn{1}{p{3.04em}}{NeRF\newline{}\makecell[c]{\cite{mildenhall2020nerf}}} & \multicolumn{1}{p{3.04em}}{NeRV\newline{}\makecell[c]{\cite{chen2021nerv}}} & \multicolumn{1}{p{3.04em}}{COIN\newline{}\makecell[c]{\cite{dupont2021coin}}} &  \makecell*[c]{MIP} & MiNL(Ours) \bigstrut\\
    		\hline
    		Model parameters↓ & 200kB & 200kB & 211kB & 218kB & 185kB & 199kB & \textbf{185kB} \bigstrut[t]\\
    		PSNR/dB ↑ & 27.12 & 26.92 & 25.23 & 22.83 & 22.88 & 23.81 & \textbf{27.71} \\
    		SSIM ↑ & 0.852 & 0.854 & 0.811 & 0.784 & 0.794 & 0.772 & \textbf{0.875} \\
    		Decoding time ↓ & 360s  & 420s  & 290s  & \textbf{1.2s}  & 187s  & 305s  & \textbf{2.2s} \bigstrut[b]\\
    		\hline
    		Model parameters↓ & 299kB & 299kB & 260kB & 284kB & 268kB & 271kB & \textbf{270kB} \bigstrut[t]\\
    		PSNR/dB ↑ & 27.89 & 27.8  & 25.76 & 26.4  & 25.63 & 25.06 & \textbf{28.89} \\
    		SSIM ↑ & 0.875 & 0.869 & 0.823 & 0.821 & 0.801 & 0.785 & \textbf{0.891} \\
    		Decoding time ↓ & 364s  & 427s  & 298s  &\textbf{ 1.6s}  & 196s  & 320s  & \textbf{2.4s} \bigstrut[b]\\
    		\hline
    		Model parameters↓ & 385kB & 385kB & 363kB & 384kB & 355kB & 360kB & \textbf{357kB} \bigstrut[t]\\
    		PSNR/dB ↑ & 28.45 & 28.51 & 26.51 & 28.08 & 26.12 & 26.03 & \textbf{29.3} \\
    		SSIM ↑ & 0.891 & 0.888 & 0.847 & 0.861 & 0.849 & 0.804 & \textbf{0.901} \\
    		Decoding time ↓ & 371s  & 434s  & 303s  & \textbf{1.9s}  & 203s  & 328s  & \textbf{2.8s}  \bigstrut[b]\\
    		\hline
    	\end{tabular}%
    	\label{tab:addlabel}%
    \end{table}%
    \section{Experimental results}
	\subsection{Implementation details}
	In this paper, the ICME 2016 Grand challenge test dataset that has 12 light field images are selected to prove the effectiveness of our proposed method. The angular resolution and spatial resolution for each light filed image are 15$\times$15 and 625$\times$434 respectively. To avoid using the dark views associated to vignetting, we perform experiments on a subset of the light field images (\textit{i.e.} central  11$\times$11 views).
	
	In our experiments, several neural networks with different model sizes are optimized for each light field image. We use the Adam optimizer with a a learning rate begins at 1$\times$$10^{-2}$ and decays gradually to 1$\times$$10^{-4}$ over the course of optimization (other hyperparameters of the Adam optimizer are set to default values). The hyperparameter $L$ of Equation \ref{pe} is set to 40, which means that the dimension of the embeeded high-dimensional input vector for MiNL is 160. The balance factor $\alpha$ of Equation \ref{loss} is set to 0.01. We train the neural network for 250 epoches, with batchsize of 5000. We evaluate the performance with two metrics: PSNR and SSIM. All experiments are run on NVIDIA GeForce RTX 3080 Laptop GPU, plaease refer to the Appendix for more experimental results.
	
	\subsection{Comparison with other implicit neural representations}
	We compare our proposed method, \textit{i.e.} MiNL, with other common implicit neural representation methods for light fields on the ICME dataset. Specifically, COIN\cite{dupont2021coin} is the most naive one where a direct mapping from raw 4D spatial-angular coordinate to pixel colors is constructed, NeRF\cite{mildenhall2020nerf} is an improvement over COIN where the Position Encoding operation is introduced, SIREN\cite{sitzmann2020implicit} and SIGNET\cite{feng2021signet} can be viewed as the state-of-the-art method that has the best visual quality nowadays. In addition to these SAI-based pixel-wise implicit representation, we also compare MiNL with an image-wise implicit neural representation (\textit{i.e.} NeRV\cite{chen2021nerv}). Morever, to validate the necessity of constructing a MI-wise implicit representation for lenslet images, we perform experiments on pixel-wise implicit representation for lenslet image (\textit{i.e.} MIP) as described in Figure \ref{fig 3} (a) and section 3.1.
	
	For fair comparison, all theses methods have the same training time and the dimension of the input coordinates of these methods are the same. As is demonstrated in Table \ref{table1} , under the condition of the same model parameters, MiNL can outperform all the other methods on PSNR, SSIM and decoding time to a large extent, with about \textbf{1$\sim$4dB} PSNR improvement and \textbf{$\times$80$\sim$180} decoding time speed-up. It is worth noting that though NeRV can have a slightly faster decoding speed than our proposed method, its visual quality is quite poor, especially when the model parameters are small.
	
	\subsection{Light field compression}
	\begin{figure}
		\centering 
		\hspace{-0.7cm}\includegraphics[height=5.5cm]{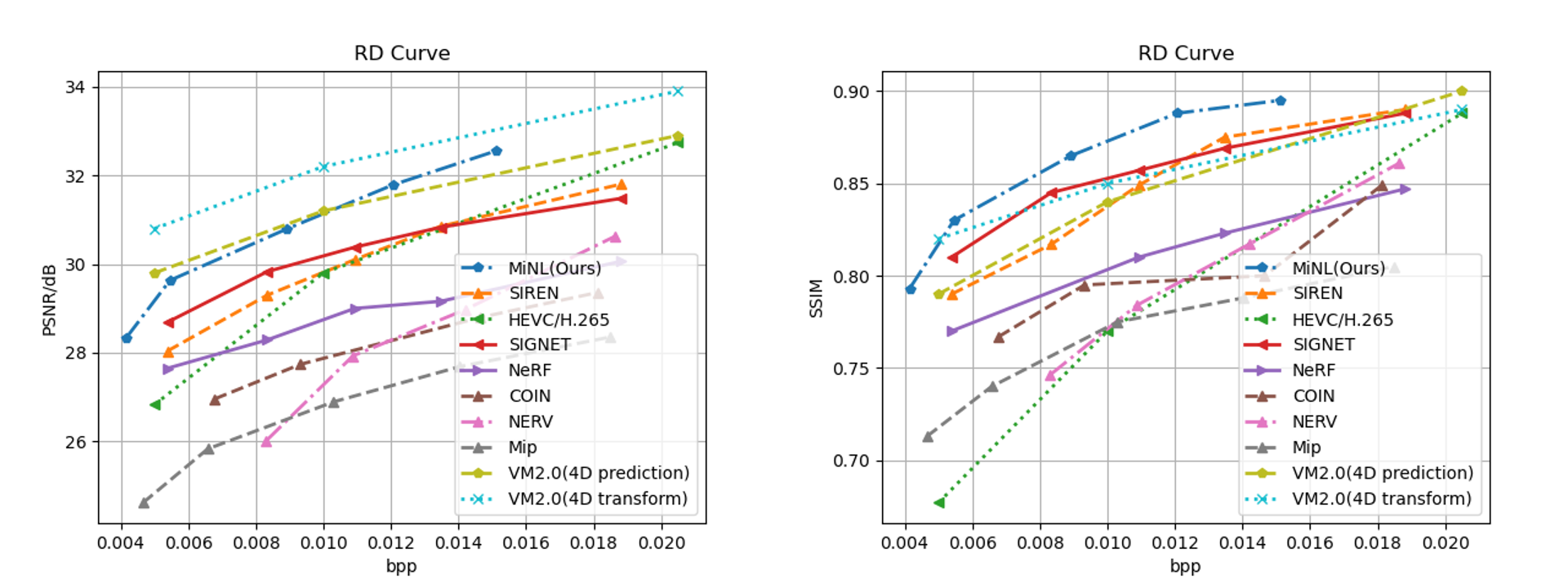}
		\caption{Rate-distortion curves for light field compression on PSNR and SSIM}\label{Compression performance}
	\end{figure}

	\begin{figure}
		\centering 
		\includegraphics[height=7.3cm]{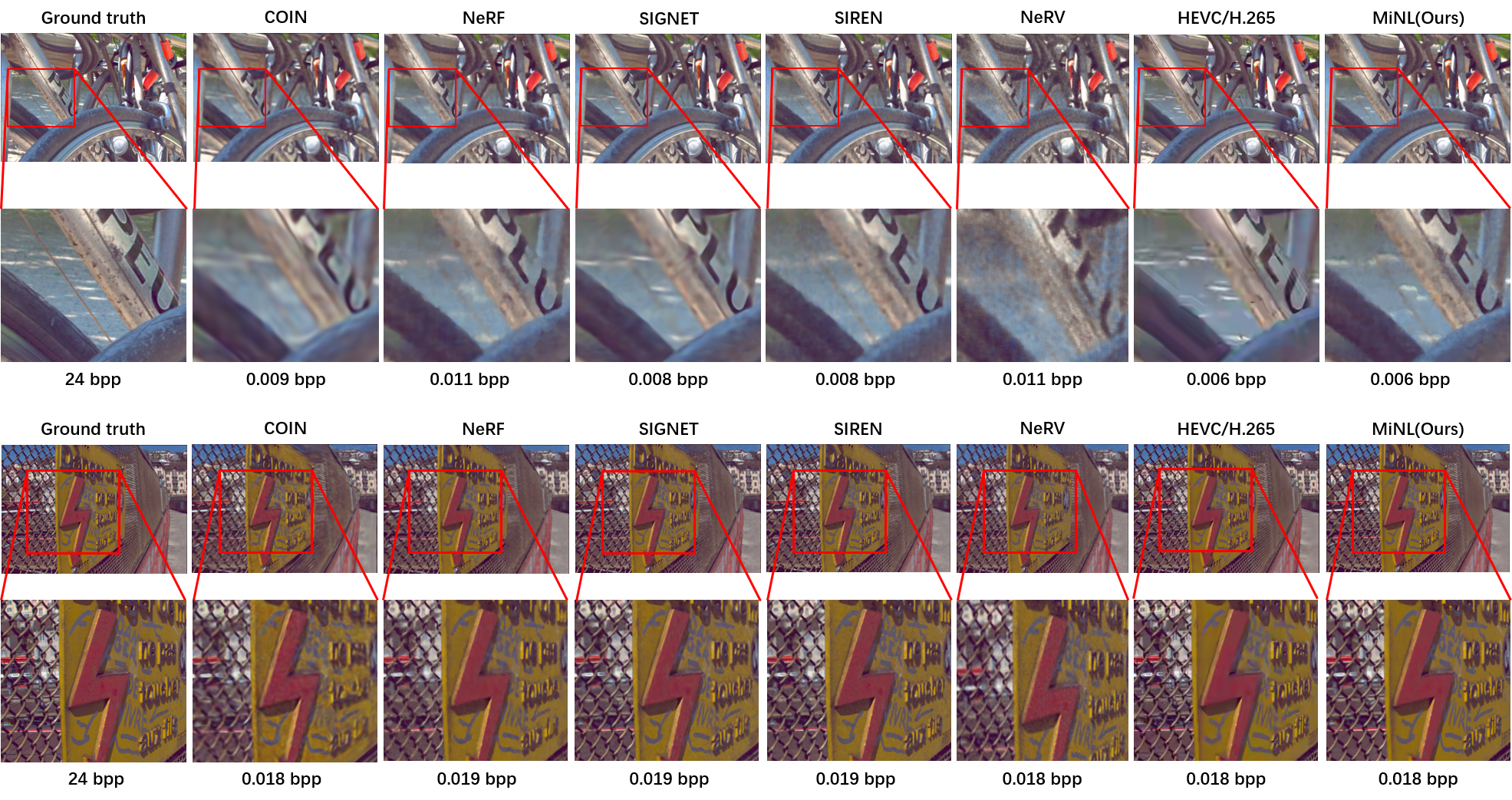}
		\caption{Visualization of light field compression for different methods}\label{Subjective comparison}
	\end{figure}
\begin{figure}[htbp]
	\centering
	\vspace{-0.1cm}
	\begin{minipage}{0.5\linewidth}
		\centering
		\captionof{table}{PSNR resulus for light field denoising. 'Baseline' means original noisy images}
		\begin{tabular}{ccccc}
			\hline
			Noise & White & Pulse & Speckle & Average \bigstrut\\
			\hline
			Baseline & 24.63 & 24.71 & 25.03 & 24.79 \bigstrut\\
			\hline
			Average & 30.22 & 30.29 & 30.18 & 30.23 \bigstrut[t]\\
			Median & 30.21 & \textbf{35.53} & 29.18 & 31.64 \\
			Gaussian & 28.25 & 28.34 & 28.54 & 28.37 \\
			MiNL  & \textbf{33.25} & 32.81 & \textbf{32.41} & \textbf{32.82} \bigstrut[b]\\
			\hline
		\end{tabular}\vspace{1cm}
		\label{tabel denoising}
	\end{minipage} 
	\hfill
	\begin{minipage}{0.45\linewidth}
		\centering
		\setlength{\abovecaptionskip}{0.1cm}
		\includegraphics[width=6cm]{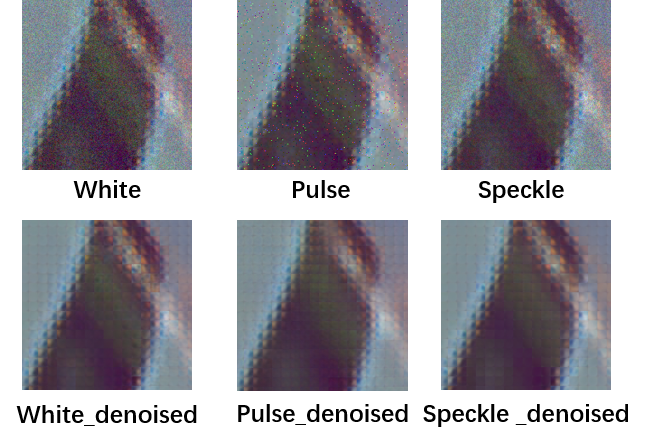}
		\caption{Visualization of denoising results for different kinds of noises}
		\label{fig denoising}
	\end{minipage}
\end{figure}
	\paragraph{Model compression results} As mentioned above, MiNL is able to transfrom the light field compression task into model compression task. To achieve the best compression performance, we firstly perform experiments on model pruning and model quantization to decide the optimal pruning ratio and quantized bit. As proved in Figure \ref{Quantization and pruning} , the model of 8 bit quantization and 20\% pruning rate can achieve almost the same performance with the orginal model with 32 bit quantization and no pruning, which we choose as the defalut setting for model compression. Entroy coding is a lossless coding method that can futher reduce the model size by 10\% $\sim$15\%.

	\paragraph{Light field compression results} We compare our proposed method with other implicit representation methods, HEVC/H.265 (x.265) \cite{pereira2019jpeg}, JPEG Pleno 4D Predictive coding mode and JPEG Pleno 4D Transform coding mode \cite{perra2019performance} for light field compression. Among these methods, HEVC is a hand-crafted coding method that treats all the light filed images as pseudo video sequences, VM 2.0(4D prediction) and VM 2.0(4D transform) are methods with quite complex processing pipelines that are specially designed for light field compression by JPEG. Bit-per-pixel (bpp) is adopted to represent the compression ratio. Figure \ref{Compression performance} shows the rate-distortion curves for different methods, where our proposed method can achieve comparable performance with the state-of-the-art method with a simple neural network training (1$\sim$2 dB PSNR improvement over HEVC and the best SSIM performance). Figure \ref{Subjective comparison} shows the visualization of light field compression results for different methods, where our proposed method can recover more high-frequency detailed information. It is worth mentioning that we only use quite simple model compression methods in MiNL, and we believe that a better model compression method will lead to a better light field compression performance.
	
    \subsection{Light field denoising}
    For light field denosing, as shown in Table \ref{tabel denoising} and Figure \ref{fig denoising} , MiNL can achieve more than 7dB PSNR improvement over the original noisy light field images by means of a simple neural network training, surpassing many traditional hand-crafted filters. It is worth noting that we only perform experiments on a model of fixed size, and if the model size grows bigger, the denosing effects of MiNL would be better.

   \section{Conslusion}
    In this paper, we present a novel micro-images based neural representation for light fields, \textit{i.e} MiNL. Different from most pixel-wise or image-wise implicit neural representations that take advantage of sub-aperture images to represent light fields, we are the first that leverage micro-images to realize MI-wise implicit representation for light fields. MiNL stores the whole light fields into parameters of neural networks and is able to realize several light field related tasks such as light field compression and light field denoising. Extensive experiments have demontrated the superiority of our proposed method, \textit{i.e.} low compliexity and higher representation ability. Compared with common implicit neural representations nowadays, MiNL can obtain 1$\sim$4dB PSNR improvement and 80$\sim$180x faster decoding speed. For light field compression, MiNL transforms it into model compression task and can achieve state-of-the-art performance compared with other implict representation methods and hand-crafted coding method such as HEVC/H.265. For light field denoising, MiNL can obtain the denoised light field images through a simple neural network training, with more than 7dB PSNR improvement over the original noisy images. The main limitation of our proposed method is that MiNL is not a generalizable method, which means that different neural networks need to be trained to overfit different light field images. Although generalization is a common problem in most implicit neural representation methods, we believe that methods such as incorporating geometry prior would help solve this problem, which will be our future work.

	\bibliographystyle{plain}
	
	\bibliography{mybib}
	

	\clearpage
	\appendix

	\section{Appendix}
	
	\subsection{Results on different light field images}
	We perform experiments on different light field images (\textit{i.e.} Danger\_de\_Mort, Stone\_Pillars\_Outside and Fountain\&Vincent2) to validate the superiority of MiNL, as demonstrated by the following results on light field compression.
	\begin{figure}[H]
		\centering 
		\includegraphics[height=3.5cm]{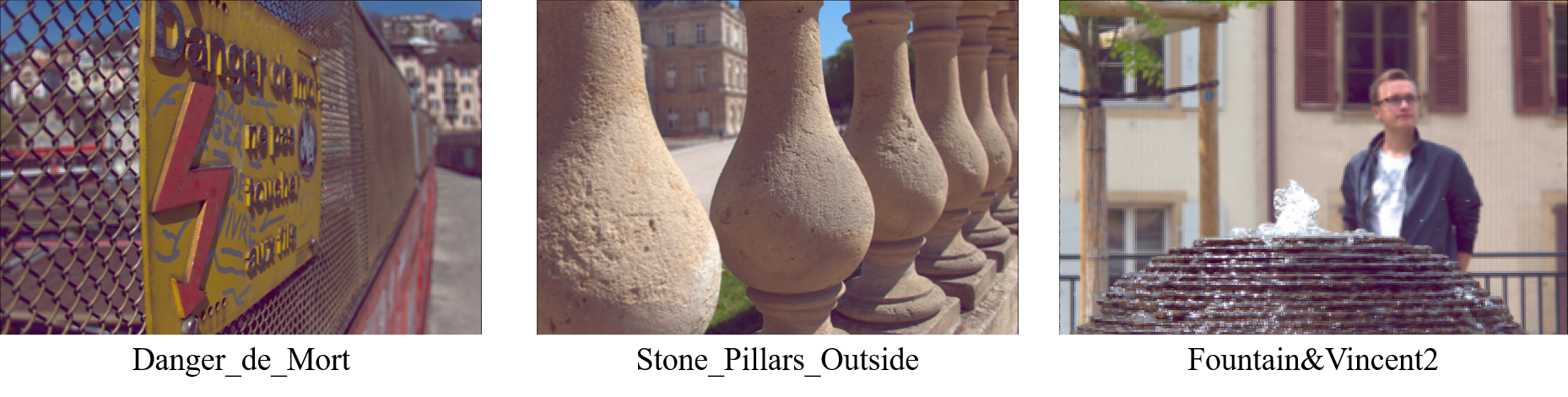}
		\caption{Different light field images used in our experiments}\label{appendix_fig1}
	\end{figure}
	
	\textbf{Danger\_de\_Mort}
	\begin{figure}[H]
		\centering 
		\hspace{-0.7cm}\includegraphics[height=5.5cm]{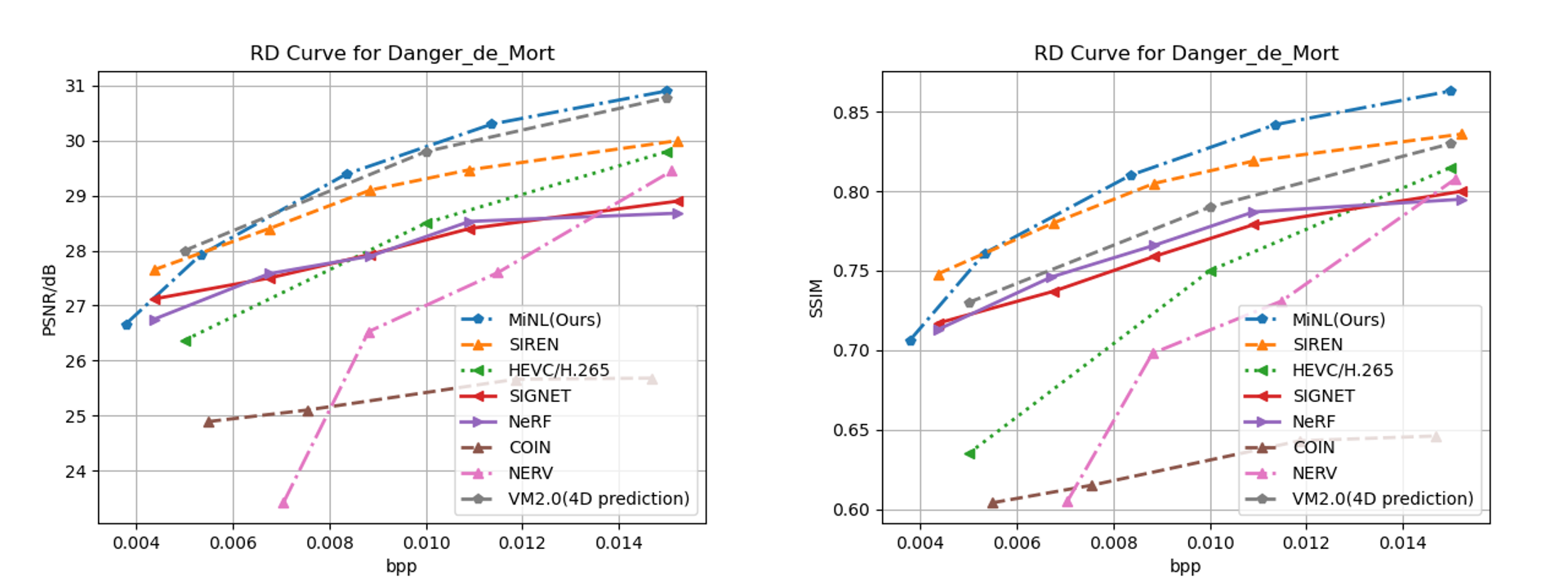}
		\caption{Light field compression results of different methods on Danger\_de\_Mort}\label{appendix_fig2}
	\end{figure}
    
    \begin{figure}[H]
    	\centering 
    	\hspace{-0.7cm}\includegraphics[height=6.5cm]{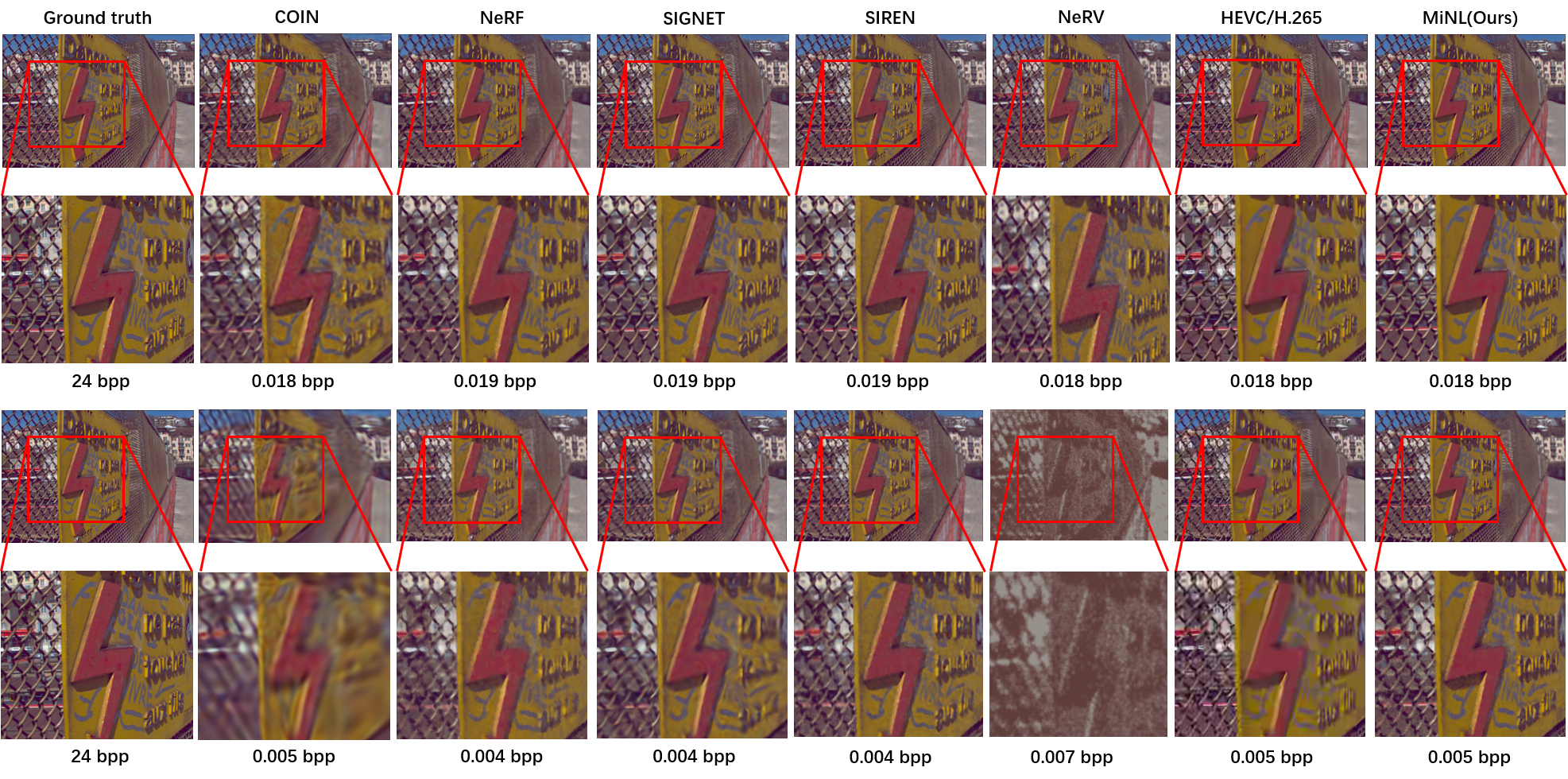}
    	\caption{Visualization of light field compression results of different methods on Danger\_de\_Mort}\label{appendix_fig3}
    \end{figure}
	
	\textbf{Stone\_Pillars\_Outside}
	\begin{figure}[H]
		\centering 
		\includegraphics[height=5.5cm]{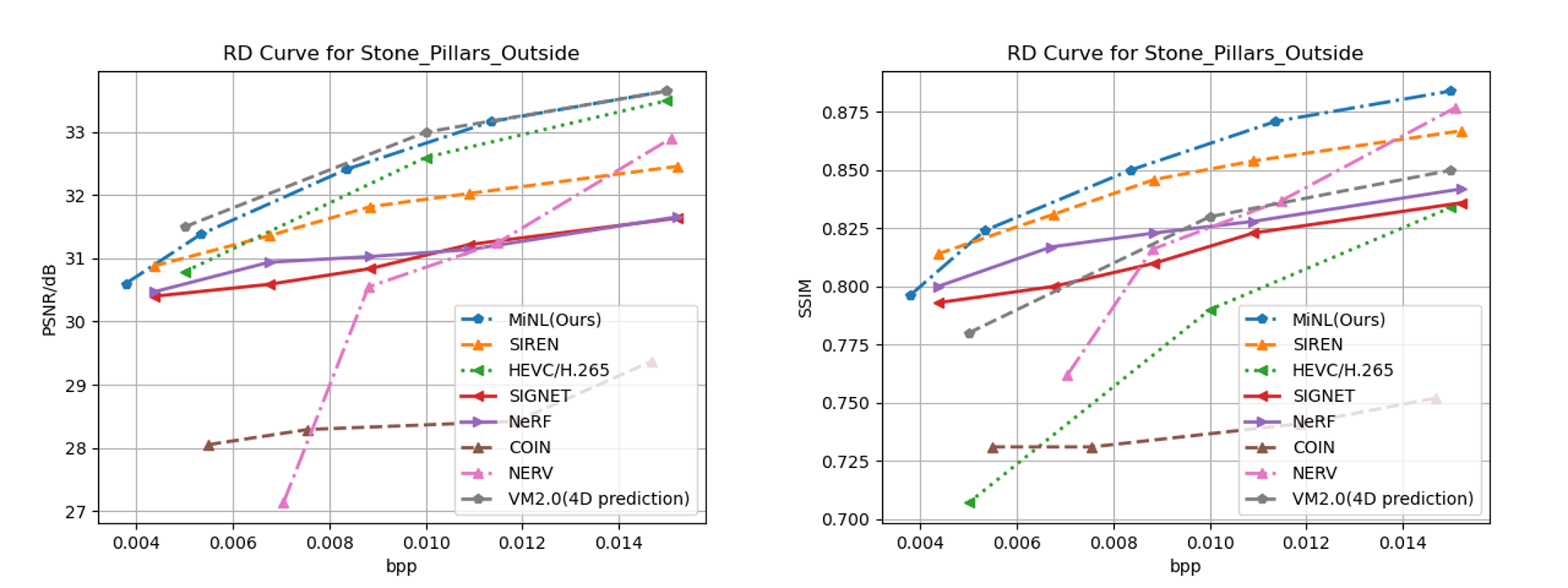}
		\caption{Light field compression results of different methods on Stone\_Pillars\_Outside}\label{appendix_fig4}
	\end{figure}
    
    \begin{figure}[H]
    	\centering 
    	\includegraphics[height=6cm]{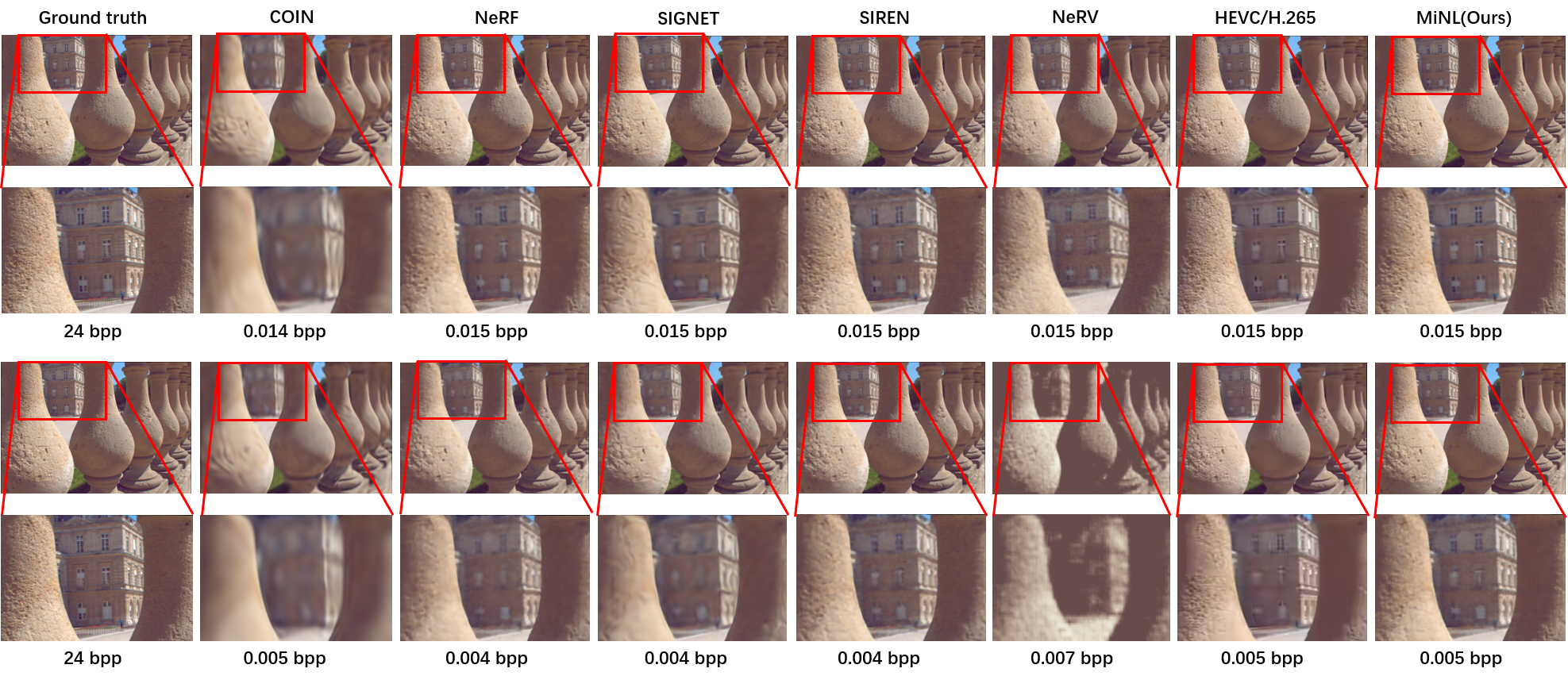}
    	\caption{Visualization of light field compression results of different methods on Stone\_Pillars\_Outside}\label{appendix_fig5}
    \end{figure}
	
	\textbf{Fountain\&Vincent2}
	\begin{figure}[H]
		\centering 
		\includegraphics[height=5.5cm]{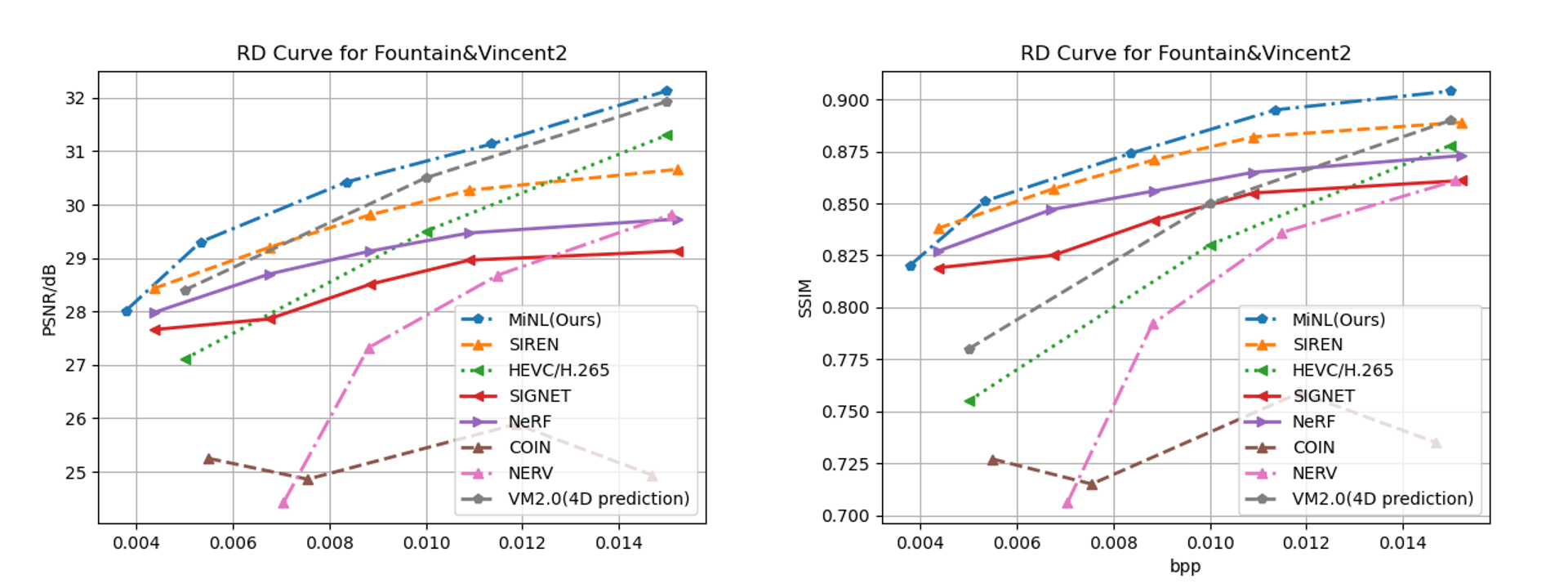}
		\caption{Light field compression results of different methods on Fountain\&Vincent2}\label{appendix_fig6}
	\end{figure}

    \begin{figure}[H]
    	\centering 
    	\includegraphics[height=6.5cm]{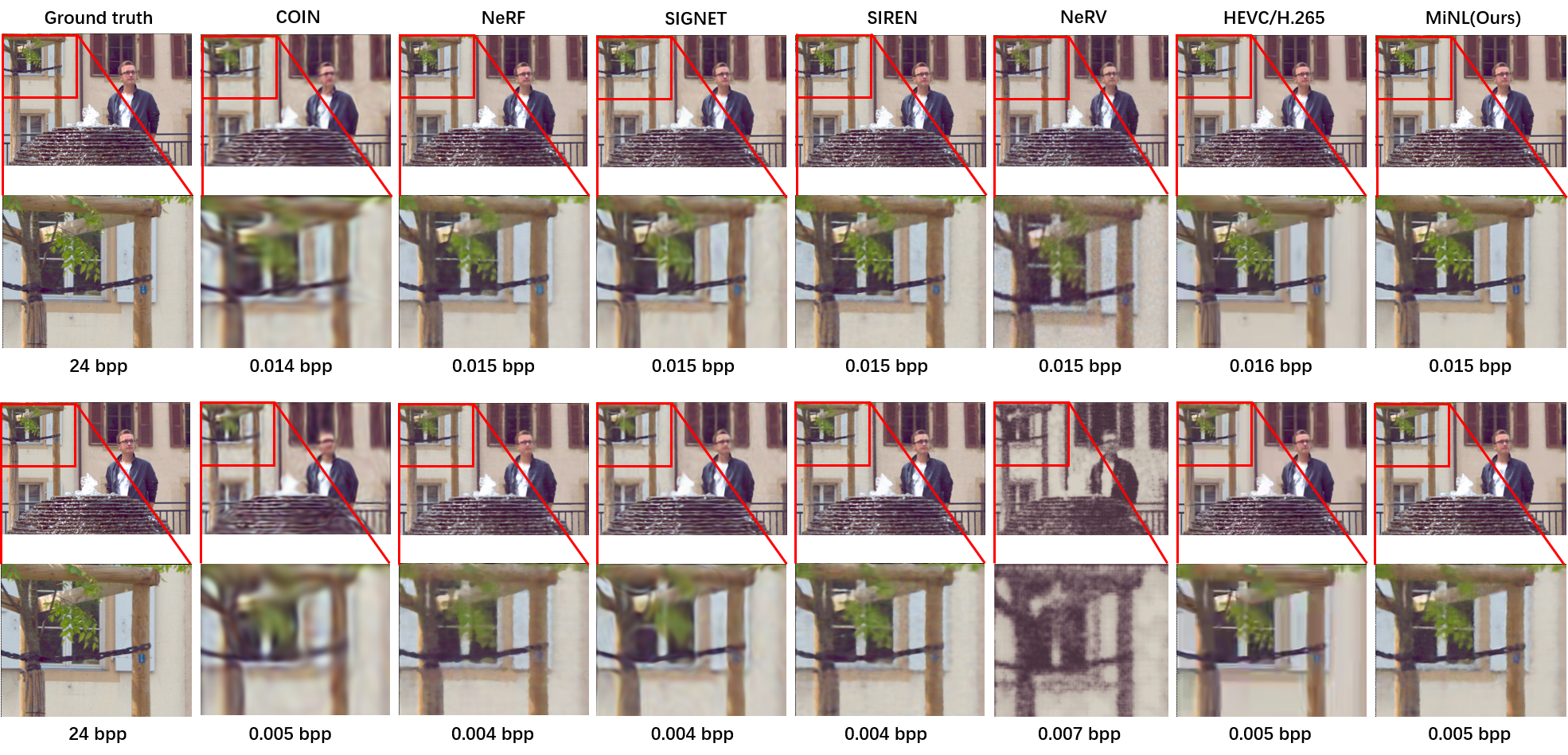}
    	\caption{Visualization of light field compression results of different methods on Fountain\&Vincent2}\label{appendix_fig7}
    \end{figure}

\end{document}